\documentclass{article}
\usepackage{authblk}
\usepackage{geometry}                % See geometry.pdf to learn the layout options. There are lots.
\usepackage{graphicx}
\usepackage{amssymb}
\usepackage{epstopdf}
\usepackage[utf8]{inputenc}
\usepackage[T1]{fontenc}
\usepackage{listings}
\usepackage{pdflscape}
\usepackage{endnotes}
\let\footnote=\endnote
\usepackage{biblatex}
\usepackage{color}
\definecolor{gray}{rgb}{0.4,0.4,0.4}
\definecolor{darkblue}{rgb}{0.0,0.0,0.6}
\definecolor{cyan}{rgb}{0.0,0.6,0.6}

\lstdefinelanguage{XML}
{
  morestring=[b]",
  morestring=[s]{>}{<},
  morecomment=[s]{<?}{?>},
  stringstyle=\color{black},
  identifierstyle=\color{darkblue},
  keywordstyle=\color{cyan},
  morekeywords={debate,page,utterance}
}

\title{PTPARL-D: Annotated Corpus of 44 years of Portuguese Parliament debates}

\date{}                                        % Activate to display a given date or no date

\begin{document}
%\maketitle

\author[1,2]{Paulo Almeida}
\author[1,3]{Manuel Marques-Pita}
\author[1,2]{Joana Gonçalves-Sá \thanks{joana.sa@novasbe.pt}}

\affil[1]{Instituto Gulbenkian de Ciência, Oeiras, Portugal}
\affil[2]{Nova School of Business and Economics, Universidade Nova de Lisboa, Carcavelos, Portugal}
\affil[3]{CICANT, Universidade Lusófona (ULHT). Lisboa, Portugal}
\maketitle
\section*{Abstract}

In a representative democracy, some decide in the name of the rest, and these elected officials are commonly gathered in public assemblies, such as parliaments, where they discuss policies, legislate, and vote on fundamental initiatives. A core aspect of such democratic processes are the plenary debates, where important public discussions take place. Many parliaments around the world are increasingly keeping the transcripts of such debates, and other parliamentary data, in digital formats accessible to the public, increasing transparency and accountability. Furthermore, some parliaments are bringing old paper transcripts to semi-structured digital formats. However, these records are often only provided as raw text or even as images, with little to no annotation, and inconsistent formats, making them difficult to analyze and study, reducing both transparency and public reach. Here, we present PTPARL-D, an annotated corpus of debates in the Portuguese Parliament, from 1976 to 2019, covering the entire period of Portuguese democracy.

\section*{Introduction}

Analyses of large numbers of parliamentary debates are intended to pave the way for a better understanding of the collective information processes involved in the approval or rejection of law proposals, agenda setting, political cycles, etc. Such analyses are also necessary to assess, question and improve internal dynamics, increasing transparency and accountability, while also facilitating comparative studies \cite{http://doi.org/10.3233/ip-2006-0095, doi:10.1080/02606755.2009.9522293}.
Traditionally, interesting political science questions on parliamentary dynamics were approached either qualitatively or, when quantitatively, focusing on very specific measurable items. Scientists would have to either focus on quantity, and give up on content analysis, or focus on content, and place human limits on how much to include. For example, longitudinal or large scale important studies performed on voting discipline and behaviour, typically do not compare between discourses, focusing on the more tractable vote count \cite{hix_noury_roland_2005, doi:10.1111/1475-6765.00106, 9781107402690}.

The increasing digitization of parliamentary documents and development of computational methods opens exciting possibilities for the fast and unbiased study of political data. Using Machine Learning for the processing of Natural Language in political texts is conceptually simple and has yielded some very interesting results \cite{HopKin10}. Therefore, starting from a text, it is possible to computationally unravel differences between discourses by analyzing the individual words (either by simply counting or ranking), sequences of words or sentence structure.
Several tools have been developed to automatically analyze complex and large political texts, which increasingly include not only manifestos but also parliamentary addresses, debates and commentary \cite{grimmer_stewart_2013, WordsasData, laver_benoit_garry_2003}, and these techniques have proven themselves, mainly focusing on the much simpler two-party US system. 

The Portuguese Parliament (PP) corpus that we now present offers not only a resource that can be used for Natural Language Processing and other text mining goals, but also a very good opportunity to study 40 years of a democratic multi-party system, including both quantitative indicators and the possibility of discourse analysis, as we describe below.
%The European Parliament (EP) also makes available all voting records and debates but, to our knowledge, no similar study of the differences between discourses of the different European MPs has been published. 

The Portuguese Parliament (PP) comprises  a single chamber -- known as  the \emph{Assembleia da República}. This is  the representative assembly of all Portuguese citizens, also overseeing the life cycles of the laws implemented in the Republic, while ensuring their compliance with the National Constitution. Portugal has records of parliamentary debates since 1821, when the country was still under monarchic ruling, through the First Republic (1911-1926), `New State' (1935-1974), and the current democratic history from 1975 to the present day. These include a number of distinct document series produced by the Portuguese government to keep records of parliamentary activities -- each corresponding to the aforementioned historic periods, -- and protocols for data recording. 

The study of the PP offers challenges both at the political science and computational levels, representing an excellent case study to approach multiparty systems. From its peaceful revolution in 1974, to joining the EU, Portugal has had some very rich 40 years of political history, while maintaining a democratic regime. 
Thus, the PP can become a great model for computational political science research, while also allowing for comparative studies, for several reasons. Firstly, it is a relatively small parliament, with the 230 elected Members of Parliament (MPs) representing the country as a whole, not a particular regional constituency and with a limited number of players: Portugal has had less than 1900 MPs since 1976, and more than 200 of those were, at some point, Members of Government (MG). Secondly, it is a multiparty system, with varying levels of polarization, that does not favour voting freedom, but allows for other types of dissent \cite{http://doi.org/10.12660/riel.v1.n1.2010.4124}; Voting against the party line is very costly and infrequent, with divergence happening mostly at the level of discourse. Thirdly, refining existing computational tools, able to detect more subtle differences, will open new possibilities to the study of other types of discourse and settings. Fourthly, because the construction of a political discourse corpus in Portuguese is a novel contribution and it is fundamental to start developing tools in important spoken languages other than English. In fact, in what concerns the debates of the PP, at the start of this project there was a corpus released as a a subset of the Corpus of Reference of Contemporary Portuguese \cite{10.1007/978-3-642-28885-2_13}, and community efforts to facilitate access to publicly available data, such as the Demo.cratica website \cite{demo.cratica}, but not a comprehensive publication of all the diaries, in a (semi-)structured format. In 2019, POPaD was launched with similar goals, but using a different methodology for speaker identification and assignment \cite{POPaD}.

The specific data source we turned into an annotated corpus is the \emph{Series I} of the `Third Republic', which stores the debate transcripts that took place in the Portuguese National Assembly since 1976, covering the full story of the Portuguese democracy. Thus, this corpus builds on the growing collection of parliamentary corpora at the European level \cite{DutchParl, OGRODNICZUK12.653, 11858/00-097C-0000-0005-CF9C-4, ABERCROMBIE18.9, NANNI18.6, WISSIK18.2, 20.500.12115/8} and offers tools for discourse analysis and comparative studies at the national and international levels.

\section*{Original Data}

Debate transcriptions were downloaded from the Portuguese Parliament (PP) website\footnote{ \texttt{(http://debates.parlamento.pt)}}. In the 8th legislature, first legislative session, the text documents from numbers 16 to 91 are empty at that location. Their content was retrieved from PDF files at the main Parliament web site\footnote{ \texttt{(http://www.parlamento.pt)}} and extracted to a similar format as the rest of the dataset. The documents have been typed by humans over a period of over forty years and, in earlier legislatures, digitized via optical character recognition (OCR). Each debate is stored as a text document. Within \emph{Series I}, the overall \emph{natural} structure of debate documents is as follows:

\begin{description}
\item[Preamble.] This includes the journal number, debate date, president chairing the debate, and secretaries.
\item[Summary.] Short statement of key topics discussed in the debate.
\item[Call of MPs.] The Parliament President calls the Members of Parliament in the assembly, assessing their presence or absence.
\item[Communications.] The President communicates relevant information for the debate.
\item[Debate transcription.] The actual body of the debate. This is the only section we annotated.
\item[Record of late MP entries.] Attendance to debate sessions is controlled, including a list of MPs who entered late.
\item[MPs that missed the debate.] List of MPs that did not attend the debate, often with a justification.
\item[Corrections/Errata.] This section contains explicit corrections to errors concerning events during the debate.
\item[Voting record.] Registry of votes by present MPs to pass or fail a discussed proposal.
\end{description}

MP names and other biographic information were also retrieved from the PP website, when available, and then stored in a relational database. This MP information is used for the assignment of speaker utterances to a specific MP during a period in which a generic \emph{`current speaker'} (`Orador') was used (more details below). In addition, for all other utterances, a match MP was identified in the database to identify the utterance's speaker unambiguously. The Parliament's website has an Open Data section with biographies of MPs, which were downloaded in the XML format. In addition, for the first five legislatures, we used CSV files provided by the Parliament administrative services.

\section{Annotation Pipeline}

Here, we are only concerned with the debate transcripts but, in order to implement our annotation scheme, we had to overcome a number of important hurdles, which can be broadly classified in three categories: 1) Identifying the beginning and ending of the actual debate; 2) identifying the beginning and ending of an utterance; and 3) identifying the MP that speaks in each utterance.

The collection of debate documents was subjected to several processing steps:

\begin{enumerate}
\item{{\bf Strip HTML}.} HTML tags were removed to leave only the text
\item{{\bf Clean headers}.} Page headers, which include page numbers and other metadata, were detected and eliminated.
\item{{\bf Clean asides}.} Asides that are not utterances, such as applause, protest and laughter, were detected and eliminated.
\item{{\bf Detect session end}.} Debate transcriptions include a section after the President of the Assembly closes the session; this is generally marked by an expression indicating the time (e.g. "Eram 18 horas", Portuguese for "It was 6 PM"), and we marked that position to eliminate the text that comes after.
\item{{\bf Tag utterances}} Utterances by individuals are generally indicated with specific punctuation marks, a colon and a dash separated by one space:

 \texttt{ <Speaker> : - <Utterance>}

We used a regular expression to detect this pattern and slight variations, to enclose all utterances in XML tags, keeping both the content (\textless Utterance\textgreater) and the person speaking (\textless Speaker\textgreater).

\item{{\bf Assign Parliament President utterances}.} President utterances are important because they can identify speakers, when the floor is granted to them. We used fuzzy matching of a few common speaker patterns (`O Sr. Presidente', `A Sr.a Presidente') to assign these utterances.

\item{{\bf Assign `Orador'}} Up until the 10th legislature, whenever an MP was interrupted, the next utterances by the same MP were marked as `O Orador', or `A Oradora' (Portuguese for `Speaker', in the masculine and feminine forms, respectively). We used simple heuristics to assign these `Orador' speakers to the names of the corresponding MPs.

\item{{\bf Assign speaker}} The speaker of each utterance was, whenever possible, identified as a specific person (i.e. an entity that can be tracked throughout the entirety of the debates, rather than just a name). Previous steps in the pipeline (tagging utterances and assigning `Orador') yielded a string representing the speaker of each utterance; this string may belong to an MP (it starts with `O Sr.' or `A Sra.', followed by the name and party), a member of Government (it contains the cabinet name) or to other actors, such as the president and secretaries of the Assembly, heads of state and guests. In the first two cases, the string was fuzzy matched against a set of MPs or members of Government, respectively, who participated in the session.

\end{enumerate}

The above steps were written as Python functions and coordinated using Luigi\footnote{ \texttt{(https://github.com/spotify/luigi)}}, a Python module to build pipelines of batch jobs. The end result of the pipeline is an XML file, whose format is described next.

\section*{Annotation Scheme}

The general annotation structure of our corpus is exemplified below:

\begin{lstlisting}
<debate period="r3" legislature="1" session="1" number="1" date="1976-06-03">
    <page number="1">
        <utterance page-start="1" speaker-string="O Sr. Presidente" speaker-role="president" order="1">Dou a palavra ao senhor Alberto Alves</utterance>
        <utterance page-start="1" speaker-id="123" speaker-name="Alberto Alves" speaker-party="AB" speaker-string="O Sr. Alberto Alves (AB)" order="2">Blah</utterance>
    </page>
</debate>
\end{lstlisting}

The outermost tag is the \textbf{$<$debate$>$}. It contains data and other substructures that pertain to a single, specific meeting of the parliament. Thus the tag includes the following fixed metadata: 

\begin{description}
\item[period] Corresponds to the historic data source, in this case always `r3', the third Republic.
\item[legislature] Indicates the government cycle in which the debate took place.
\item[legislative\_session] Typically refers to a year period within each 4-year legislature.
\item[number] Ordinal, per legislature session index of the debate.
\item[date] Date when the parliamentary session took place (not the publication date of the document, which is usually one or more days later)
\end{description}

Within the \textbf{$<$debate$>$} tag, each page is enclosed in a \textbf{$<$page$>$} sub-tag that corresponds to the page number within the published journal containing the debate. This information is kept in the annotation scheme to help with the identification of utterances in the source journals. Finally, within each page there is one or more $<$utterance$>$ sub-tags, containing the words spoken by MPs and other actors in the Parliament debate session. Utterances contain several attributes: 

\begin{description}
\item[page-start.] Is the page number where the utterance begins.
\item[speaker-string] The original string identifying the speaker, from which the other speaker-* attributes are derived. We keep this in the document because it is useful when the speaker is not (correctly) identified by the pipeline.
\item[speaker-id.] This is a universal ID that identifies the MP uniquely across the entire corpus.
\item[speaker-name.] Short name of the speaker, speaker-id will support any possible need for disambiguation.
\item[speaker-party.] This is the party the speaker belonged to when the utterance was produced.
\item[speaker-role.] Is currently only used for utterances by the President of the National Assembly, where its value is `president'. Other utterances do not have this attribute.
\item[order.] Is the ordinal number specifying the order of the utterance in the current debate.
\end{description}

\section*{Corpus Description}

The corpus comprises 4448 debates, divided into 13 legislatures (see Table \ref{tab:stats_table}), each of which may contain up to four legislative sessions. The average number of utterances per debate is 305.12 (with median 273) and the standard deviation is 194.38. The maximum number of utterances in a single debate is 2104. There are, on average, 95.20 words per utterance (median of 19 and standard deviation of 198.30), with the longest utterance having 16558 words.

\pagestyle{empty}
\begin{landscape}
\begin{table}
\begin{tabular}{c|c|c|c|c|c}
    \hline
    Legislature & Start Date & End Date & Number of debates & Average utterances per debate & Average words per utterance  \\ \hline
    1 &	1976-06-02 & 1980-07-14	& 416 & 319.98 & 96.06 \\
    2 &	1980-11-13 & 1983-05-30 & 289 & 323.77 & 96.01 \\
    3 &	1983-05-31 & 1985-11-03 & 269 & 347.24 & 94.27 \\
    4 &	1985-11-04 & 1987-08-12	& 191 & 255.86 & 119.35 \\
    5 &	1987-08-13 & 1991-11-03	& 449 & 281.49 & 105.02 \\
    6 &	1991-11-04 & 1995-10-26	& 393 & 243.54 & 110.56 \\
    7 &	1995-10-27 & 1999-10-24	& 411 & 316.41 & 92.09 \\
    8 & 1999-10-25 & 2002-04-04	& 230 & 352.61 & 86.67 \\
    9 &	2002-04-05 & 2005-03-09	& 278 & 398.23 & 78.54 \\
    10 & 2005-03-10 & 2009-10-14 & 472 & 329.67 & 80.12 \\
    11 & 2009-10-15 & 2011-06-19 & 157 & 299.76 & 89.18 \\
    12 & 2011-06-20 & 2015-10-22 & 475 & 283.77 & 86.55 \\
    13 & 2015-10-23 & 2019-10-24 & 417 & 255.25 & 92.62 \\
    \hline
\end{tabular}    
    \caption{Start and end dates for each legislature, along with the corresponding number of debates, average number of utterances per debate and average number of words per utterance.}
    \label{tab:stats_table}
\end{table}

\end{landscape}
\pagestyle{plain}
\section*{Acknowledgments}
The authors would like to thank Pedro Varela and Manuel Arriaga for early help with scraping; Demo.cratica, especially Ricardo Lafuente and Ana Isabel Carvalho, for sharing code and information on retrieving data from the Parliament's web site; Manuela Magalhães who provided important information on the disambiguation of the speakers and on their biographies; and members of the DS\&P lab at NOVA SBE for critical reading of the manuscript.

\section*{Funding}
This work was partially supported by PTDC IVC ESCT 5337 2012, funded by the Portuguese Fundação para a Ciência e para a Tecnologia (FCT) and by the Welcome DFRH WIIA 60 2011, co-funded by the FCT and the Marie Curie Actions, both awarded to JGS. The funders had no role in study design, data collection and analysis, decision to publish, or preparation of the manuscript.
\theendnotes
\printbibliography

\end{document}